\documentclass[runningheads]{llncs}
\usepackage{graphicx}
\usepackage{amsmath}
\usepackage{textcomp}
\usepackage{listings}
\usepackage{xcolor}
\usepackage{hyperref}
\usepackage{multirow}

\colorlet{punct}{red!60!black}
\definecolor{background}{HTML}{EEEEEE}
\definecolor{delim}{RGB}{20,105,176}
\colorlet{numb}{magenta!60!black}

\lstdefinelanguage{json}{
    basicstyle=\scriptsize\ttfamily,
    numbers=none,     captionpos=b,
    numberstyle=\scriptsize,
    stepnumber=1,
    numbersep=8pt,
    showstringspaces=false,
    breaklines=true,
    frame=lines,
    backgroundcolor=\color{background},
    literate=
     *{0}{{{\color{numb}0}}}{1}
      {1}{{{\color{numb}1}}}{1}
      {2}{{{\color{numb}2}}}{1}
      {3}{{{\color{numb}3}}}{1}
      {4}{{{\color{numb}4}}}{1}
      {5}{{{\color{numb}5}}}{1}
      {6}{{{\color{numb}6}}}{1}
      {7}{{{\color{numb}7}}}{1}
      {8}{{{\color{numb}8}}}{1}
      {9}{{{\color{numb}9}}}{1}
      {:}{{{\color{punct}{:}}}}{1}
      {,}{{{\color{punct}{,}}}}{1}
      {\{}{{{\color{delim}{\{}}}}{1}
      {\}}{{{\color{delim}{\}}}}}{1}
      {[}{{{\color{delim}{[}}}}{1}
      {]}{{{\color{delim}{]}}}}{1},
}

\begin{document}
\title{\textit{ElasticHash}: Semantic Image Similarity Search by Deep Hashing with Elasticsearch}
\titlerunning{\textit{ElasticHash}}
\author{Nikolaus Korfhage
\and
Markus Mühling
\and
Bernd Freisleben}
\authorrunning{Korfhage et al.}
\institute{Department of Mathematics and Computer Science,\\ University of Marburg, Marburg, Germany\\
\email{\{korfhage,muehling,freisleb\}@informatik.uni-marburg.de}\\
}
\maketitle              \begin{abstract}
We present \textit{ElasticHash}, a novel approach for high-quality, efficient, and large-scale semantic image similarity search. It is based on a deep hashing model to learn hash codes for fine-grained image similarity search in natural images and a two-stage method for efficiently searching binary hash codes using Elasticsearch (ES). 
In the first stage, a coarse search based on short hash codes is performed using multi-index hashing and ES terms lookup of neighboring hash codes. In the second stage, the list of results is re-ranked by computing the Hamming distance on long hash codes.
We evaluate the retrieval performance 
of \textit{ElasticHash} for more than 120,000 query images on about 6.9 million database images of the OpenImages data set. The results show that our approach achieves high-quality retrieval results and low search latencies.
\end{abstract}

\section{Introduction}
Query-by-content approaches based on feature representations that are learned by deep convolutional neural networks (CNNs) have greatly increased the performance of content-based image retrieval systems.
However, state-of-the-art methods in the field of semantic image similarity search suffer from shallow network architectures and small data sets with few image classes in the training as well as in the evaluation phases.  Few image classes in the training phase lead to poor generalizability to query images with unknown content in the evaluation phase, i.e., a more fine-grained modeling of the image content is required. Thus, high accuracy for arbitrary search queries, fast response times, and scalability to millions of images are necessary to meet many users' needs both in scientific and commercial applications. 

In this paper, we present \textit{ElasticHash}, a high-quality, efficient, and scalable approach for semantic image similarity search based on the most popular enterprise full-text search and analytics engine Elasticsearch\footnote{\url{https://www.elastic.co}} (ES).
ES processes queries very fast due to inverted indices based on Lucene\footnote{\url{https://lucene.apache.org}}, scales to hundreds of servers, provides load balancing, and supports availability and reliability. Apparently, the properties of ES are not only desirable for full-text search, but also for semantic image similarity search. Furthermore, integrating image similarity search into ES allows multi-modal queries, e.g., combining text and images in a single query. 
The contributions of the paper are as follows:
\begin{itemize}
    \item We present \textit{ElasticHash}, a novel two-stage approach for semantic image similarity search based on multi-index hashing and integrate it via terms lookup queries into ES.
                \item  We present experimental results to show that \textit{ElasticHash} achieves fast response times and high-quality retrieval results at the same time by leveraging the benefits of short hash codes (better search times) and long hash codes (higher retrieval quality). To the best of our knowledge, we provide the first evaluation of image similarity search for more than 120,000 query images on about 6.9 million database images of the OpenImages data set.
    \item We make our deep image similarity search model, the corresponding ES indices, and a demo application available
        at \url{http://github.com/umr-ds/ElasticHash}.             \end{itemize}

The paper is organized as follows. In Section \ref{relatedwork}, we discuss related work. Section \ref{search} presents \textit{ElasticHash}. In Section \ref{evaluation}, we evaluate \textit{ElasticHash} on the OpenImages data set in terms of search latency and retrieval quality. Section \ref{conclusion} concludes the paper and outlines areas for future work.

\section{Related Work}
\label{relatedwork}
Deep learning, in particular deep CNNs, led to strong improvements in content-based image similarity search. With increasing sizes of the underlying image databases, the need for an efficient similarity search strategy arises. Since high-dimensional CNN features are not suitable to efficiently search in very large databases, large-scale image similarity search systems focus on binary image codes for quantization or compact representations and fast comparisons rather than full CNN features. 

Recently, several deep hashing methods were introduced \cite{zhu2016deep,erin2015deep,wang2015learning,wang2018survey,cao2017binary,liu2016deep,cao2018deep}. Many of them employ pairwise or triplet losses. While these methods often achieve state-of-the-art performance on their test data sets, they are not necessarily suitable for very large data sets and fine-grained image similarity search based on thousands of classes. Existing deep hashing methods are often trained using small CNNs that usually cannot capture the granularity of very large image data sets. Often, CNN models like AlexNet \cite{krizhevsky2012imagenet} are used as their backbones, 
and they are usually evaluated on a small number of image classes \cite{wang2012semi,cao2018deep,erin2015deep,zhu2016deep} (e.g., a sample of 100 ImageNet categories \cite{cao2018deep}, about 80  object categories in COCO \cite{lin2014microsoft}, NUS-WIDE \cite{chua2009nus} with 81 concepts, or even only 10 classes as in MNIST or CIFAR). Additionally, the image dimensions in CIFAR and MNIST are very small (32x32 and 28x28, respectively) and thus not sufficient for image similarity search in real-world applications. Many approaches are trained on relatively small training data sets (e.g., 10,000  - 50,000  images \cite{cao2017hashnet,cao2018deep,liu2016deep}). In addition, there are no standardized benchmark data sets, and each publication uses different splits of training, query, and database images, which further complicates a comparison of the methods. Furthermore, training from scratch can be prohibitively expensive for large data sets. We observed that for large data sets with a high number of image classes, a transfer learning approach that combines triplet loss and classification loss leads to good retrieval results. 
To the best of our knowledge, \textit{ElasticHash} is the first work that presents a deep hashing model trained and evaluated on a sufficiently large number of image classes. 

The currently best performing approaches for learning to hash image representations belong either to product quantization (PQ) methods \cite{jegou2010product,johnson2019billion} and methods based on deep hashing (DH) \cite{wang2015learning,erin2015deep}. 
Amato et al. \cite{AmatoLargeScaleImageRetrievalwithElasticsearch} present PQ approaches that transform neural network features into text formats suitable for being indexed in ES. However, this approach cannot match the retrieval performance of FAISS \cite{johnson2019billion}.
Therefore, we focus on deep hashing that in combination with multi-index hashing (MIH) \cite{norouzi2012fast} can circumvent exhaustive search in Hamming space and achieve low search latency while maintaining high retrieval quality.

\textit{ElasticHash} is related to other image similarity search methods integrated into ES. For example, FENSHSES \cite{mu2019fast} integrates MIH into ES and has a search latency comparable to FAISS. The method works efficiently for small radii of the Hamming ball and relatively small data sets (500,000 images). Small hamming radii, however, often produce too few neighbors for a query \cite{norouzi2012fast}. MIH like FENSHSES is thus not suitable for our scenario of large-scale image retrieval in ES with long binary codes (256 bits), where we require sub-second search latency on a data set of about 7 million images. Furthermore, we solve the shortcomings of FENSHSES using only a subset of bits rather than the whole hash codes to perform our MIH-based coarse search. While other works extend ES for image similarity search by modifying the Lucene library \cite{Gennaro2010}, our approach is seamlessly integrated into ES without modifying its code base.

\section{\textit{ElasticHash}} 
\label{search}
\begin{figure}[tb]
\centerline{\includegraphics[width=1\textwidth]{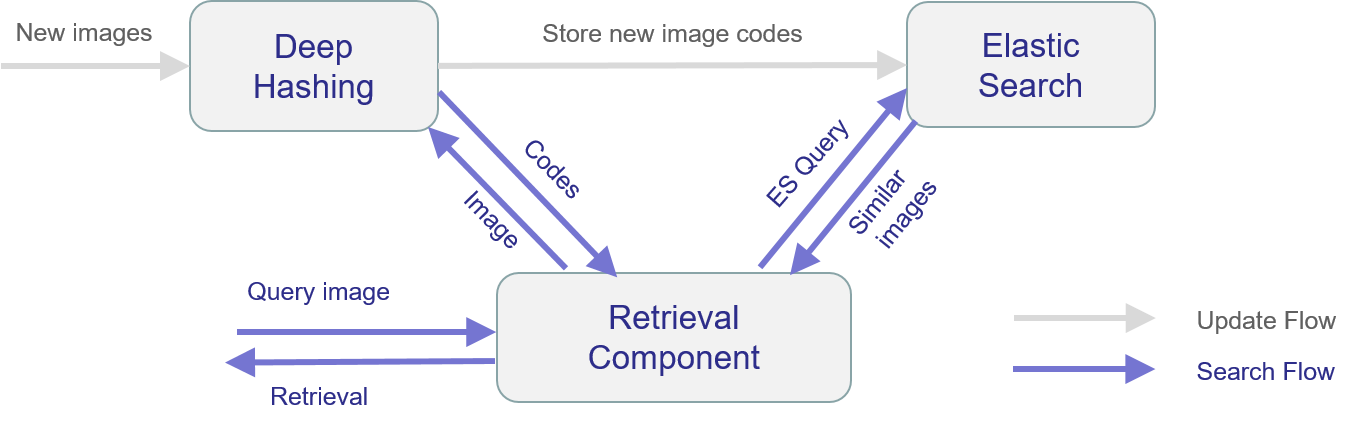}}
\caption{Overview of the workflows for image similarity search in ES.}
\label{overview}
\end{figure}

\textit{ElasticHash} consists of several components as shown in Figure \ref{overview}: a deep hashing component, an ES cluster, and a retrieval component. The deep hashing component is realized as a web service using Tensorflow Serving where the integrated deep hashing model is applied to images and the corresponding binary codes are returned. 
In the first phase, the binary codes are extracted from the database images in the indexing phase using the deep hashing component and stored into the ES cluster. After initially building the index, the retrieval component handles incoming query images and visualizes the retrieval results.
For this purpose, the binary codes are extracted from the query images using the web service, the corresponding ES queries are assembled and sent to the ES cluster that returns the final list of similar images.
The entire similarity search system can be easily deployed for production via Docker.

The deep hashing model is described in more detail in Section \ref{subsec:model}, including the training strategy and network architecture. In Section \ref{subsec:es_integration}, the ES integration is presented.

\subsection{Deep Hashing Model}
\label{subsec:model}
We now describe our deep hashing model and how it is used to extract both short and long hash codes.

The model training consists of two phases that both use ADAM as the optimization method. First, an ImageNet-pretrained EfficientNetB3 \cite{tan2019efficientnet} model is trained on a data set with a larger number of classes in order to obtain a more fine-grained embedding. In contrast to the original ImageNet dataset, it contains all ImageNet classes with more than 1000 training images and all classes of the Places2 \cite{zhou2017places} data set, which results in a total number of classes of 5,390. The model is trained with cross-entropy loss on a Softmax output. After two epochs of training the final layer with a learning rate of 0.01, all layers are trained for another 16 epochs with a learning rate of 0.0001. 

In the second phase, the classification model's weights are used to initialize the deep hashing model. This model includes an additional 256-bit coding layer before the class output layer with $tanh$ activation and 256 outputs.  This model is trained for 5 epochs with a learning rate of 0.0001. It is trained on the same data set as before, however, by combining cross-entropy loss on the output and hard triplet loss \cite{schroff2015facenet} on the coding layer. 

With the classification loss 
\begin{equation}
\mathcal{L}_{c} = \sum_{i=1}^{K} y_i \log{p}_i
\end{equation} for $K$ classes with labels $y_i$ and predictions $p_i$, and the triplet loss 
\begin{equation}
\mathcal{L}_{t} =  max(d(a, p) - d(a, n) + \gamma, 0) 
\end{equation} for Euclidean distance $d$ between the 256-dimensional output of the coding layer for anchor image $a$ and positive example $p$ and between $a$ and a negative example $n$, respectively, the combined loss function is given by:
\begin{equation}
\mathcal{L} =  \alpha \mathcal{L_{c}} + \beta \mathcal{L_{t}},
\end{equation}
where we set margin $\gamma = 2$ and weights  $\alpha= 1$ and $\beta = 5$. We first sample a batch of size $b=128$ images from a uniform distribution of the classes. This batch is used for both computing the classification loss and generating $b$ hard triplets.
To make the similarity search more robust, we used heavy data augmentation in both phases, which in addition to standard augmentation methods includes inducing JPEG compression artifacts. 
After training, the model generates 256-bit codes. These codes can be decomposed into four 64-bit codes for fast computation of Hamming distance on long integers. However, using codes of this length on a corpus of about 10 million images is too expensive, even when using multi-index hashing. We therefore extracted 64-bit codes from the original 256-bit codes to perform the filtering on shorter codes and thus smaller Hamming ball radii. To extract the 64 most important bits from the 256-bit codes, we first partition the 256-bit codes into four partitions by applying the Kernighan-Lin algorithm \cite{kernighanlin} on the bit correlations. From each of the four decorrelated partitions, we then take the first 16 bits to compose 64-bit codes.

\subsection{Integration into ES}
\label{subsec:es_integration}
Before describing our image similarity search integration into ES, we will shortly review MIH in Hamming space \cite{norouzi2012fast}.
The idea of MIH is based on the following observation: for two binary codes $h=(h^1,...,h^m)$ and $g=(g^1,...,g^m)$ where $m$ is the number of partitions, $h^k$ and $g^k$ are the $k^{th}$ subcodes and $H$ is the Hamming norm, the following proposition holds:
\begin{equation}\left \| h - g \right \|_H \leq r  \Rightarrow
\exists k \in \left \{ 1,...,m \right \} \: \left \| h^k - g^k \right \|_H \leq \left \lfloor \frac{r}{m} \right \rfloor 
\label{prop}
\end{equation}

For the case of 64-bit codes that are decomposed into $m=4$ subcodes, this means that a code is in a Hamming radius $r < 12$ if at least one of the subcodes has a distance of $d \leq \lfloor \frac{r}{m} \rfloor = 2$ from the query subcode. The performance of MIH can be increased if the subcodes are maximally independent of each other \cite{WanDatadrivenmultiindexhashing}, especially for shorter codes \cite{mu2019fast}. Thus, after training a deep hashing model, the bit positions should be permutated accordingly.

The ES index used for retrieval contains four short codes (\texttt{f\_0 - f\_3}) and four long subcodes (\texttt{r\_0 - r\_3}) for each image.
The short codes are used for MIH and efficiently utilize the reverse index structure of ES and are thus separated into four subcodes of type "keyword". The long codes are also separated into four subcodes in order to allow fast computation of Hamming distances for values of type long.

An additional index is used for fast lookup of neighboring subcodes within the retrieval query. The neighbors index does not change once it has been created and merely serves as an auxiliary index for term queries. It requires pre-computing all nearest neighbors for all possible 16-bit subcodes. Thus, the index of neighbors contains $2^{16}$ documents. The document id corresponds to the unsigned integer representation of a 16-bit subcode and can therefore accessed within a term query. It contains a single field "nbs" that is assigned to a list of all neighboring 16-bit codes within a Hamming radius of $d$ of the corresponding query subcode. Since this index basically works as a lookup table, it could also be realized somewhere else, i.e., not as an ES index. However, integrating the lookup table this way eliminates the need for external code and enables fast deployment of the whole system.  
All documents representing all possible 16-bit subcodes are inserted according to the query in Listing \ref{lst:es-nbs}.

\label{lst:nbs}
\begin{lstlisting}[caption={Query for adding an entry to neighbor lookup index.},label=lst:es-nbs,language=json,firstnumber=1,]
POST /nbs-d2/_doc/<16 bit subcode>
{ "nbs" : [ <d2 neighbors of 16 bit subcode>  ] }
\end{lstlisting}

In this stage, MIH is realized by querying the additional index of neighbors for fast neighbor lookup. Even with MIH, using the full code length of the deep hashing model trained for 256-bit codes is too expensive for larger databases. We therefore limit the code length for the filtering stage to 64-bit codes.
To obtain a sufficiently large set of candidate hash codes in the first stage, we need to search within a Hamming ball with a correspondingly large radius. We set $d=2$, which will return at least all codes within $r=11$ of a 64-bit code. In our setting with $d=2$, this results in $137$ neighbors per subcode, i.e., $548$ neighbors in total. 

In ES, we realize MIH by using a terms lookup. It fetches the field values of an existing document and then uses these values as search terms (see Listing \ref{lst:es-nbs}). In contrast to putting all neighbors into the query, using a dedicated index for subcode neighbors has the advantage that the retrieval of neighboring subcodes is carried out within ES. Thus, the query load is small, and no external handling of neighbor lookup is necessary.

In the second stage, all codes obtained by MIH are re-ranked according to their Hamming distance to the long code. To compute the Hamming distance of the 256-bit code, the Painless Script in Listing \ref{lst:es-script} is applied to each of the four subcodes.

\begin{lstlisting}[caption={Query for adding a Painless Script.},label=lst:es-script,language=json,firstnumber=1,]
POST _scripts/hd64
{ "script": { "lang": "painless", 
  "source": 64-Long.bitCount(params.subcode^doc[params.field].value) } }
\end{lstlisting}

The query in Listing \ref{lst:es-query} combines the MIH step as a filter with a term query and the re-ranking step as an application of the painless script from Listing \ref{lst:es-script} on the filtered retrieval list.

\begin{lstlisting}[caption={Query for performing two-stage similarity search.},label=lst:es-query,language=json,firstnumber=1,float,floatplacement=H]
GET /es-retrieval/_search
{ "query": {
"function_score": {
 "boost_mode": "sum",  "score_mode": "sum",
 "functions": [ ..., {
  "script_score": {
   "script": { "id": "hd64",
    "params": {
     "field": "r_<i>",
     "subcode":  <64 bit subcode for re-ranking> } } }, "weight": 1 }, ... ],
"query": {
  "constant_score": {
   "boost": 0.
   "filter": {
    "bool": {
     "minimum_should_match": 1,
     "should": [ ..., {
      "terms": {
       "f_<j>": {
        "id": "<16 bit subcode for lookup>",
        "index": "nbs-d2",
        "path": "nbs" } } }, ... ] } } } }, } } }
\end{lstlisting}

\section{Experimental Evaluation} 
\label{evaluation}
To determine the search latency and retrieval quality of \textit{ElasticHash}, we evaluate three settings for using the binary hash codes generated by our deep hashing model for large-scale image retrieval in ES: (1) short codes, i.e., 64 bits for both filtering and re-ranking, (2) long codes, i.e., 256 bits for both filtering and re-ranking, and (3) \textit{ElasticHash}, i.e., 64 bits for filtering, 256 bits for re-ranking.
Settings (1) and (2) are similar to the MIH integration of Mu et al. \cite{mu2019fast}. 

To evaluate our approach, we use OpenImages \cite{kuznetsova2020open}, which is currently the largest annotated image data set publicly available. It contains multi-label annotations for 9.2 million Flickr images with 19,794 different labels and is partitioned into training, validation, and test data set. On the average, there are 2.4 positive labels for the training split, while the validation and test splits have 8.8.
As our database images we use all training images being available when downloading the data set, i.e., 6,942,071 images in total.
To evaluate the retrieval quality, we use all downloaded images from the OpenImages test and validation set as query images (121,588 images in total). From these images, we draw a sample of 10,000 images to measure the search latencies for the three different settings.

The quality of the retrieval lists is evaluated using the average precision (AP) score, which is the most commonly used quality measure in image retrieval. The AP score is calculated from the list of retrieved images as follows:
\begin{equation}
\label{eq:fund:ap}
AP(\rho)=\frac{1}{\left|R \cap \rho^N\right|}\sum_{k=1}^N \frac{\left|R \cap \rho^k\right|}{k} \psi(i_k), \\
\textrm{with} \quad \psi(i_k) =
\begin{cases} 
\ 1 & \mbox{if } i_k \in R 
\\[5pt] \ 0  & \mbox{ otherwise } 
\end{cases}
\end{equation}
where $N$ is the length of the ranked image list,
$\rho^k=\{i_1,i_2,\ldots,i_k\}$ is the ranked image list up to rank $k$, $R$
is the set of relevant documents, $\left|R \cap \rho^k\right|$ is the number of relevant images in the top-$k$ of $\rho$ and $\psi(i_k)$ is the relevance function.
We consider an image as relevant, if it has at least one label in common with the query image.
To evaluate the overall performance, the mean AP score is calculated by taking the mean value of the AP scores over all queries.

\subsection{Results}
We first evaluate the search latency for the queries. Next, we compare the retrieval quality in terms of AP.
The experiments were performed on a system with an Intel Core i7-4771 CPU @ 3.50GHz and 32 GB RAM. 

\begin{figure}[tb]
\centerline{\includegraphics[width=1\textwidth]{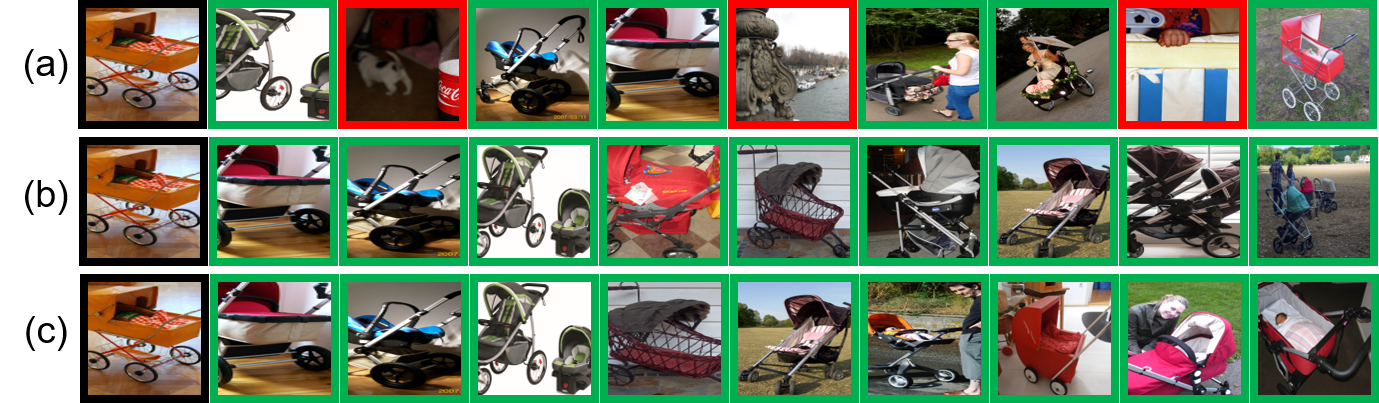}}
\caption{Top-10 retrieval results for (a) short codes, (b) long codes, and (c) \textit{ElasticHash} for the same query image (first on the left); green: relevant result; red: irrelevant result.}
\label{fig:retrieval}
\end{figure}

\begin{table}
\centering
\caption{Retrieval quality in terms of mean AP for different thresholds of $k$ on 121,588 query images.}
\begin{tabular}{|r||r|r|r|r|r|r|r|}
\hline
top $k$ & 10 & 25 & 50 & 100 & 250 & 500 & 1000 \\
\hline
\hline
short & 87.94 & 86.08 & 84.44 & 82.54 & 79.41 & 76.44 & 72.86 \\
\hline
long & 95.35 & 94.72 & 94.23 & 93.71 & 92.90 & 92.09 & 90.95 \\
\hline
\textit{ElasticHash} & 95.21 & 94.48 & 93.90 & 93.22 & 92.02 & 90.61 & 88.42 \\
\hline
\end{tabular}
\label{tab:aps}
\end{table}

\begin{table}
\centering
\caption{Search latencies for ES queries (ms) with standard deviation for different thresholds of $k$ on 10,000 query images.}
\begin{tabular}{|l||r|r|r|r|r|r|r|r|}
\hline
top $k$ & & 10 & 25 & 50 & 100 & 250 & 500 & 1000 \\
\hline
\hline
\multirow{2}{*}{short} & $\mu$ & 23.09 & 23.98 & 24.45 & 25.58 & 28.38 & 33.09 & 42.20 \\
 & $\sigma$ & 4.74 & 4.65 & 4.70 & 4.72 & 4.86 & 5.20 & 6.07 \\
\hline
\multirow{2}{*}{long} & $\mu$ & 111.83 & 111.58 & 111.99 & 113.05 & 116.77 & 121.98 & 132.60 \\
 & $\sigma$ & 16.50 & 16.58 & 16.72 & 16.54 & 17.04 & 17.13 & 17.99 \\
\hline
\multirow{2}{*}{\textit{ElasticHash}} & $\mu$ & 36.12 & 36.75 & 37.28 & 38.17 & 40.88 & 45.73 & 55.23 \\
 & $\sigma$ & 7.80 & 7.96 & 7.81 & 7.89 & 7.93 & 8.12 & 8.64 \\
\hline
\end{tabular}
\label{tab:times}
\end{table}

Table \ref{tab:aps} shows that for a $k$ up to 250 there is no notable decrease in retrieval quality when employing \textit{ElasticHash} rather than using the long codes for both stages. Figure \ref{fig:retrieval} shows examples of the top-10 retrieval results for the three settings. It is evident that the retrieval quality of \textit{ElasticHash} is similar to using long codes, and both are superior to using short codes. On the other hand, Table \ref{tab:times} indicates that the average retrieval time only slightly increases compared to using short codes for both stages. This suggests that \textit{ElasticHash} is a good trade-off between retrieval quality and search latency. 
Although our deep hashing model was trained on 5,390 classes, but almost 20,000 classes occur in the validation data set, high AP values are achieved for \textit{ElasticHash}. 

\section{Conclusion}
\label{conclusion}
We presented \textit{ElasticHash}, a novel two-stage approach for semantic image similarity search based on deep multi-index hashing and integrated via terms lookup queries into ES.  
Our experimental results on a large image data set demonstrated that we achieve low search latencies and high-quality retrieval results at the same time by leveraging the benefits of short hash codes (better search times) and long hash codes (higher retrieval quality). 

There are several areas for future work. For example,
it would be interesting to investigate how many classes are necessary to obtain a high degree of generalizability. Furthermore, our loss function could be adapted to multi-label image data.  Finally, we plan to extend our approach to achieve intentional image similarity search \cite{korfhage2020intentional} using ES.

\section{Acknowledgements}
This work is financially supported by the German Research Foundation (DFG project number 388420599) and HMWK (LOEWE research cluster Nature 4.0).

\bibliographystyle{splncs04}
\bibliography{refs}

\end{document}